\documentclass[runningheads]{llncs}
\usepackage[T1]{fontenc}
\usepackage{graphicx}
\usepackage{float}    
\usepackage{makecell}
\usepackage[hidelinks]{hyperref}
\begin{document}
\title{Data Overdose? Time for a Quadruple Shot: Knowledge Graph Construction using Enhanced Triple Extraction
\thanks{%
\textbf{Preprint (author’s original).} This is a preprint of a chapter accepted for \emph{SAICSIT 2025}, published in \emph{Communications in Computer and Information Science (CCIS), Springer}. 
The final authenticated version is available at \url{https://doi.org/10.1007/978-3-031-96262-2_15}{doi:10.1007/978-3-031-96262-2\_15}.%
}}
\titlerunning{Knowledge Graph Construction using Enhanced
Triple Extraction}
%
\author{Taine J. Elliott\inst{1}\orcidID{0009-0005-7553-1759}
\and
Stephen P. Levitt\inst{1}\orcidID{0000-0001-6054-6134} \and
Ken Nixon\inst{1}\orcidID{0000-0001-5391-8147} \and
Martin Bekker\inst{1}\orcidID{0000-0001-8766-937X}
}

\authorrunning{T. Elliott et al.}
%
\institute{School of Electrical and Information Engineering,\\
University of the Witwatersrand, Johannesburg, South Africa\\
\email{stephen.levitt@wits.ac.za, taine.elliott1@students.wits.ac.za}}
%
\maketitle              
\begin{abstract}


The rapid expansion of publicly-available medical data presents a challenge for clinicians and researchers alike, increasing the gap between the volume of scientific literature and its applications. The steady growth of studies and findings overwhelms medical professionals at large, hindering their ability to systematically review and understand the latest knowledge. This paper presents an approach to information extraction and automatic knowledge graph (KG) generation to identify and connect biomedical knowledge. Through a pipeline of large language model (LLM) agents, the system decomposes 44 PubMed abstracts into semantically meaningful proposition sentences and extracts KG triples from these sentences. The triples are enhanced using a combination of open domain and ontology-based information extraction methodologies to incorporate ontological categories. On top of this, a context variable is included during extraction to allow the triple to stand on its own -  thereby becoming `quadruples'. The extraction accuracy of the LLM is validated by comparing natural language sentences generated from the enhanced triples to the original propositions, achieving an average cosine similarity of 0.874. The similarity for generated sentences of enhanced triples were compared with generated sentences of ordinary triples showing an increase as a result of the context variable. Furthermore, this research explores the ability for LLMs to infer new relationships and connect clusters in the knowledge base of the knowledge graph. This approach leads the way to provide medical practitioners with a centralised, updated in real-time, and sustainable knowledge source, and may be the foundation of similar gains in a wide variety of fields.

\keywords{Knowledge graphs  \and Information extraction \and Biomedical knowledge graphs \and Coreference resolution \and Semantic chunking \and Context-Aware extraction.}
\end{abstract}
\section{Introduction}

The expansion of medical data poses a challenge for both clinicians and researchers, widening the gap between evidence generation and its practical application. The volume of new studies and findings, produced at an exponential rate, overwhelms the capacity of medical professionals to systematically review and apply the latest knowledge. Kamtchum-Tatuene and Zafack \cite{ref_stroke} have shown that there were more than 30,000 articles containing the term `stroke' published in 2020 alone, which is approximately five times the output of 2000 and about 2.3 times that of 2010. This volume of research makes it nearly impossible for practitioners to remain informed about the latest evidence. Furthermore, the challenge is increased by the workload associated with reviewing and synthesizing this literature; Alper et al. \cite{ref_jmla} estimate that keeping up-to-date with primary care publications would require the equivalent of two to three full-time researchers.




This study investigates the feasibility of employing a pipeline of LLM agents to decompose medical abstracts into semantically meaningful proposition sentences, extract corresponding KG triples, and enhance these triples into contextually complete quadruples. The aim is to generate a dynamic knowledge graph that not only reflects the current state of biomedical research, but also facilitates the inference of new relationships by connecting previously isolated knowledge clusters with one another. By doing so, the research demonstrates the possibility of combining the rapidly growing biomedical literature into a single unified source.


\section{Theoretical Foundations}

The recent expansion of data has created a need for systems capable of linking diverse and dispersed information. Traditional databases store data in flat formats that fail to express the relationships between different pieces of information, this tends to make it difficult to integrate and interpret information from disparate sources. Knowledge graphs address this challenge by structuring data to connect entities through relationships, forming a network that mirrors how knowledge naturally exists \cite{hogan2021kg}. For example, consider the sentence:
\begin{quote}
    ``Smoking increases the risk of Pancreatic cancer.''
\end{quote}
This sentence can be transformed into the triple in the format (subject, relationship, object):
\[
\texttt{(Smoking, increases the risk of, Pancreatic cancer)}
\]
Collectively, these triples form a comprehensive model of a knowledge domain. By interlinking individual triples, a network of entities and relationships that reflects the complex structure of real-world information is formed. Various graph data models represent this structure in different ways. For instance, directed edge-labelled graphs express relationships as labeled connections between nodes, property graphs enrich nodes and edges with additional attributes, and heterogeneous graphs explicitly distinguish between different types of entities and relationships \cite{hogan2021kg}. The following figure shows a property graph format of a KG.

\begin{figure}[H]
    \centering
    \includegraphics[width=0.8\linewidth]{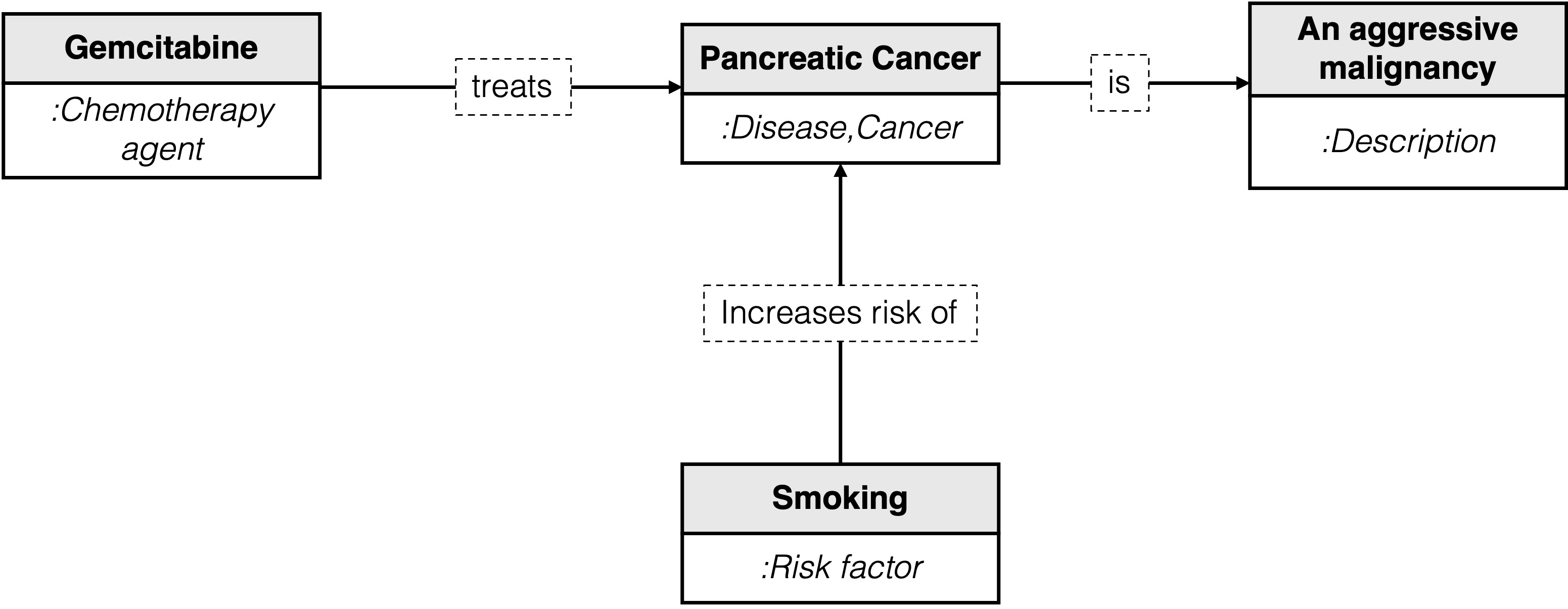}
    \caption{An example of a property knowledge graph for pancreatic cancer.}
    \label{fig:kg_example}
\end{figure}

Moreover, ontologies and schemas provide an essential layer of formalization that defines both the meaning (semantics) and the constraints of the data in a knowledge graph, thereby establishing a common vocabulary for the domain. An ontology is a formal, explicit specification of a shared conceptualisation—it defines the set of concepts (classes), relationships (properties), and rules that govern how those concepts interact within a specific domain \cite{hogan2021kg}. In other words, it lays out what entities exist and how they relate to one another. Schemas, in this context, serve as structural blueprints that describe how data should be organized within the graph, specifying the expected format of nodes and edges and the types of values they can hold \cite{hogan2021kg}. 


\section{Background}

\subsection{Problem background}

The ever-growing volume of medical knowledge is spread across many different platforms and databases rather than being stored in one central repository. As a result, researchers and clinicians must gather information from a variety of sources each using their own formats and standards, making it difficult to stay up-to-date with the latest evidence.. This situation is further complicated by the diversity of medical data, which can vary according to legal, linguistic, and technological factors \cite{DeGiacomo2020ECAI}. Current knowledge and data structures often do not represent the full complexity of a domain because traditional systems, such as relational databases, flat file repositories, and standard digital libraries, tend to treat information as isolated entries rather than as interconnected concepts \cite{Codd1970,Halevy2006}. This approach means that even when large volumes of data are available, the inherent relationships among various pieces of information are not clearly represented. Consequently, researchers and clinicians must manually piece together disparate data, which can lead to a fragmented understanding of the overall picture. Without structured semantic links that reveal how concepts connect, it is difficult to gain a holistic view of the field of research and make well-informed decisions. Although medical ontologies such as MeSH~\cite{Nelson2001} and UMLS~\cite{Bodenreider2004} do exist to link concepts and facilitate interoperability between knowledge bases, they have not proven to be a complete solution for the dispersion of information due to challenges such as limited coverage or rigid structures~\cite{Tudorache2013}. The purpose of an ontology is to provide a structured framework that defines concepts and their relationships ideally enabling a more interconnected view; however these are not universally adopted or consistently implemented across institutions. In practice, their coverage is limited, their structures overly rigid, and their integration with existing data systems fragmented ~\cite{Tudorache2013}. As a result, while they offer a promising starting point, medical ontologies alone have not been sufficient to overcome the challenges posed by the dispersion and heterogeneity of medical knowledge.


\subsection{Related work}


Information Extraction (IE) is the process of extracting structured information from unstructured text into a structured knowledge base \cite{liu2024survey}. This structured information is extracted in the form of KG triples which take the form (subject, relationship, object). Traditionally, this process involves Named Entity Recognition for identifying the subject and object entities to be extracted and then Relation Extraction (RE) for extracting the relationship between these entities. IE can be broadly categorized into two main methodologies: Ontology-based IE (OBIE) and Open Domain IE (OpenIE) \cite{liu2024survey}. OBIE systems operate under predefined schemas and taxonomies, extracting KG triples based on structured ontologies and fixed relationship templates \cite{niklaus2018survey}. These systems are guided by expert-curated biomedical ontologies such as the Unified Medical Language System (UMLS), Gene Ontology, and SNOMED CT \cite{Liu2016}. These ontologies provide a controlled vocabulary that ensures consistency in extracted entities and relations, making OBIE highly interpretable, precise, and easily transferable to knowledge bases that follow the same ontological standard \cite{Kilicoglu2009}. However, OBIE suffers several flaws: i.) The reliance on predefined schemas means that new or emerging biomedical knowledge may not be captured if it does not fit within the existing ontology \cite{Liu2016}.
ii.) It is difficult to constantly maintain and update ontologies. As biomedical knowledge grows, manually expanding ontologies to incorporate new relationships is a time-consuming and labor-intensive process \cite{Rindflesch2003}.
iii.) Differences in terminologies and conceptual structures across various biomedical ontologies can lead to inconsistencies when integrating multiple knowledge bases \cite{Kilicoglu2009}.
iv.) OBIE systems often suffer from low recall. Because they only extract information explicitly defined in the ontology, they may fail to capture novel information mentioned in biomedical literature \cite{Kilicoglu2009}.
In contrast, OpenIE exists as an alternative option, focusing on extracting knowledge triples without prior constraints on relationship or entity types \cite{banko2007open}. This process involves extracting KG triples from all sentences irrespective of their importance. This flexibility allows OpenIE to process large-scale unstructured text efficiently, making it more versatile for diverse domains of text and extracting novel information \cite{xu2023generative}. However, OpenIE systems have several flaws:
i.) Extracted entities and relations are often ambigious and inconsistent as OpenIE does not distinguish between multiple meanings of the same word across different contexts \cite{fader2011reverb}.
ii.) OpenIE systems struggle with context and complex sentences as these systems extract triples sentence by sentence \cite{fader2011reverb}.  
iii.) OpenIE systems often suffer from low precision. As the systems captures all possible KG triples in the text \cite{fader2011reverb}.
Relation extraction (RE) is a subset of Information Extraction that focuses on identifying and classifying relationships between entities in unstructured text. Given a piece of text and a set of recognized entities, RE systems determine whether a meaningful relationship exists between them and, if so, classifies its type. These relationships are represented in the form of KG triples (subject, relation, object) \cite{Jiang2022}. In the biomedical domain, RE plays a critical role in structuring knowledge from scientific literature, electronic health records, and clinical trial reports. Biomedical RE focuses on identifying meaningful associations between biological entities such as genes, proteins, diseases, drugs, and treatments. These relationships are fundamental for drug discovery, precision medicine, and disease understanding \cite{Gurulingappa2012}. Biomedical RE is particularly challenging due to the complexity of medical language, the variability in relation expression, and the need for domain-specific knowledge \cite{Liu2016}.
Transformer-based models, particularly those derived from BERT (Bidirectional Encoder Representations from Transformers), have become the foundation of modern biomedical relation extraction (RE) \cite{Vaswani2017}. Several domain-specific adaptations have further enhanced RE performance. BioBERT, trained on PubMed and biomedical literature, outperforms generic BERT models in extracting relations from medical texts \cite{Lee2020}. Similarly, SciBERT, trained on a broader scientific corpus, provides robust representations for biomedical and scientific domains \cite{Beltagy2019}. PubMedBERT, which is exclusively trained on PubMed abstracts, further refines biomedical entity and relation extraction \cite{Gu2021}. These models are typically fine-tuned using labeled biomedical RE datasets such as ChemProt, DDI, and BioCreative challenges, achieving state-of-the-art performance in structured relation classification. However, despite their success, they are an exhibit of OBIE and are only able to extract KG triples that they were trained on.
The emergence of LLMs has created a new paradigm for relation extraction. Models in the GPT family, such as GPT-3 and GPT-4, demonstrate adept ability to perform a variety of NLP tasks via prompting, even without direct fine-tuning on those tasks \cite{Brown2020}. Researchers have recently explored formulating relation extraction as a text generation problem: given an input sentence, the model is prompted to output the relation triple or a natural-language description of the relationship between entities. Hu \textit{et al.} (2023) evaluated GPT-3.5 and GPT-4 on several biomedical RE datasets in zero-shot and one-shot settings and found that it can achieve remarkably high accuracy \cite{Hu2023}. Wadhwa \textit{et al.} (2023) demonstrated that prompting GPT-3 with a few demonstrations of RE can yield performance on par with fully supervised models on standard datasets. An advantage of this generative approach is flexibility: the model is not constrained to a fixed relation ontology. However, it introduces challenges in evaluation (since the same relation can be expressed in different words) \cite{Wadhwa2023}. Mihindukulasooriya \textit{et al.} (2023) introduced TEXT2KGBENCH, a benchmark designed to evaluate the capability of LLMs in generating knowledge graphs from natural language text while strictly adhering to a given ontology. They provided two datasets each comprising thousands of sentence–triple alignments, and defined seven evaluation metrics including fact extraction accuracy, ontology conformance, and hallucination rates. Their baseline experiments used a few-shot learning prompt where the ontology was given in the context window. The results using local models showed a strong ontology conformance \cite{Text2KGBench2023}. However, a clear limitation to this approach is the limitation of context windows in LLMs and their ability to perform a task with a large number of tokens \cite{levy2024sametask}.
Medical knowledge graphs (MKGs) have emerged as a crucial tool for structuring and integrating vast amounts of biomedical knowledge. Unlike general-purpose knowledge graphs, MKGs are tailored to capture domain-specific relationships between entities such as diseases, genes, drugs, and clinical symptoms. These structured representations facilitate numerous applications, including clinical decision support, drug discovery, and personalized medicine \cite{yang2024hkgreview}. The integration of large-scale biomedical datasets from sources such as electronic health records, medical literature, and ontologies presents both opportunities and challenges in constructing MKGs. Traditional methods have relied on manual curation and rule-based approaches, but recent advancements leverage deep learning, graph-based reasoning, and LLMs to automate and enhance the construction of MKGs \cite{khalid2024mkgautomation}. 
Recent efforts in MKG automation (M-KGA) have focused on improving completeness and connectivity by employing ontology-guided enrichment and graph completion techniques. The M-KGA framework \cite{khalid2024mkgautomation} proposes a hybrid approach that integrates BioPortal ontologies with pretrained embeddings, facilitating automatic knowledge discovery from unstructured EHR data. By leveraging ClinicalBERT using traditional IE methods, M-KGA enhances entity recognition and relationship extraction, enabling a more comprehensive representation of biomedical concepts. Furthermore, cluster-based and node-based comparison methods have been introduced to uncover hidden relationships between medical entities, improving link prediction within the graph. The role of MKGs in clinical decision support is particularly significant. The review by Yang et al.(2024) highlights the utilization of MKGs in disease prediction, drug repurposing, and patient stratification \cite{yang2024hkgreview}. For instance, integrating MKGs with LLMs has enabled enhanced knowledge retrieval for clinical risk assessment, reducing reliance on manually curated databases. Additionally, ontology-driven frameworks ensure standardization and interoperability between diverse biomedical data sources, enhancing the overall quality of AI-driven medical applications. Despite these advancements, challenges such as data privacy, scalability, and real-world adoption remain active areas of research in the field of medical knowledge graphs.



\section{Information Extraction Pipeline}
This research aims to achieve the following outcomes:
i.) Create an IE pipeline that uses the capabilities of LLMs to combine both OBIE and OpenIE, thus allowing for knowledge extraction that is able to extract novel triples and conform to an ontology. 
ii.) Current approaches only consider the extraction of knowledge as triples, this research proposes the extraction of a fourth variable which allows for context based extraction.
iii.) Current approaches are ineffective at handling inferred knowledge and only extract what exists in the text. This research aims to mitigate that and provide a more connected KG.

The system is designed to take in a chunk of text, in this case an abstract. This chunk goes through a preprocessing phase where it is converted into sentences that can stand on their own, representing discrete units of knowledge, and entities in these sentences are resolved for clarity. These sentences are then sent for extraction where knowledge triples are extracted from the sentences and enriched. These enriched triples are then reconstructed into a natural language sentence so that their extraction can be validated against the original sentence. The enriched triples are then reasoned over to extract any triples that may have been inferred. The following diagram represents the systems flow.

\begin{figure}[H]
    \centering
    \includegraphics[width=1.0\linewidth]{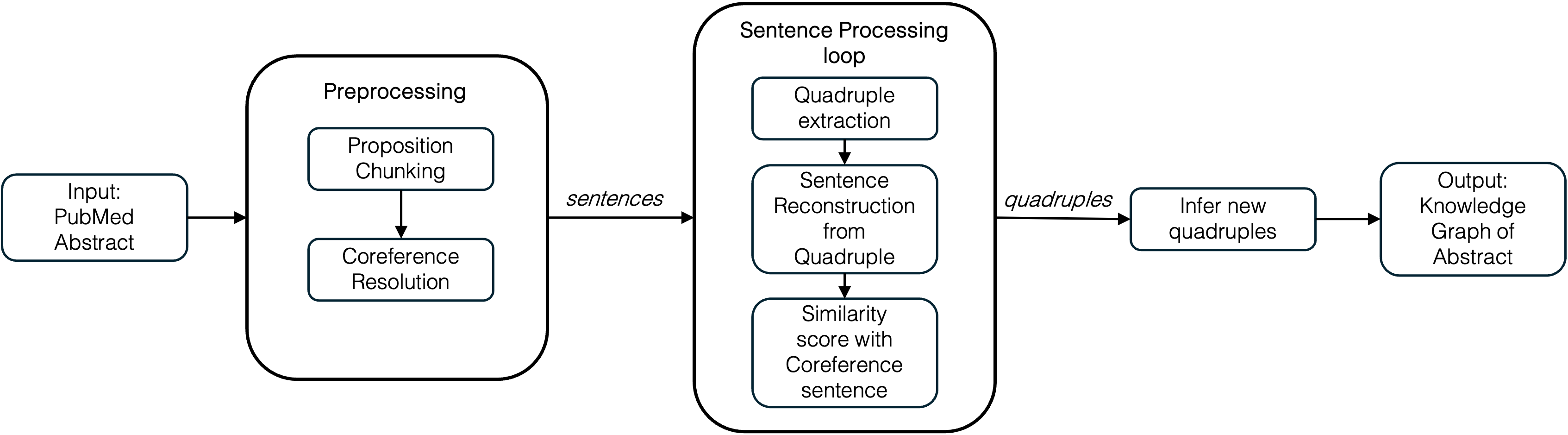}
    \caption{Overview of IE Pipeline for Automated Knowledge Graph Generation from Natural Language Text (PubMed Abstract).}
    \label{fig:system_overview}
\end{figure}


\subsection{Preprocessing - Chunking}


Levy et al. show that increasing the number of input tokens to a LLM reduces the reasoning performance of the model \cite{levy2024sametask}. Therefore, in order for an LLM to produce the most accurate information extraction, it will require the smallest semantically complete unit of text. Chunking is the process by which large amounts of text are converted into smaller, manageable segments \cite{kdb2024chunking}. Chen et al. (2024) investigate the impact of retrieval granularity (chunking) on downstream tasks, showing that whilst larger paragraphs do contain additional context, they tend to distract LLMs in downstream tasks. Whilst sentence-level chunking improves accuracy but loses context \cite{chen2024denseXretrieval,shi2023distracted}. Therefore, Chen et al. proposed propositional chunking, where a proposition is an atomic unit of knowledge that encapsulates a fact in natural language \cite{chen2024denseXretrieval}. They state that a proposition must have 3 qualities: i.) it must contain a distinct segment of meaning in the text, where the sum of all propositional meanings represents the entire semantic meaning of the text. ii.) The proposition should not be able to be split into smaller sentences. iii.) The proposition should contain all the necessary context to stand on its own and interpret what it means.
They used GPT-4 to generate propositions and performed a manual quality analysis on the propositions generated against the rules above. They found that in the 408 propositions generated: 0.7\% were not faithful to the text, 2.9\% were not split to their most minimal form and 4.9\% were not stand alone and distinct in their meaning. They tested proposition chunking with different retrievers on different datasets and found that the retrieval of proposition chunks consistently outperformed that of sentence and paragraph chunks on all datasets \cite{chen2024denseXretrieval}.

Therefore, the decision was made to use proposition-based chunking with GPT-4 because:
i.) The size of the chunks are small and contain all the relevant context, which allows for the LLM to extract the most information possible out of the input text.
ii.) GPT-4 has a low error rate (0.7\%) when extracting propositions from text. Even though 7.8\% error might account for propositions not being in their most minimal form and containing external context, it is presumed that LLMs will still be able to extract the relevant information.
iii.) Ultimately, when retrievers are used on the knowledge graph, propositions show the highest promise of returning the most relevant information.

The following diagram shows an example of a paragraph being separated into propositions.

\begin{figure}
    \centering
    \includegraphics[width=1.0\linewidth]{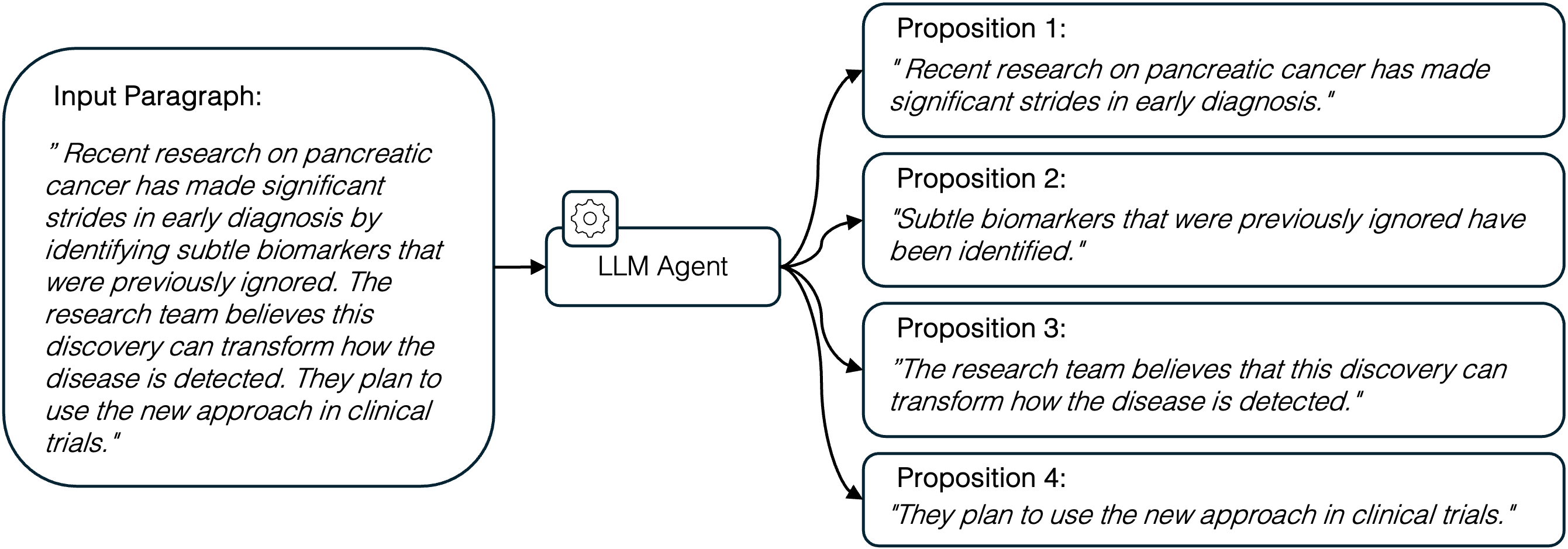}
    \caption{Example of Proposition System.}
    \label{fig:proposition_example}
\end{figure}

\subsection{Preprocessing - Co-reference Resolution}


Co-reference resolution is the task of identifying when two or more linguistic expressions refer to the same real-world entity in text \cite{Zheng2012}. In practice, this means detecting ``mentions'' of the same thing and linking them together. For example, in a clinical note the phrase “the electrocardiogram” might later be mentioned as “it”, and these must be recognized as co-referent (both referring to the same test) \cite{Zheng2012}. In biomedical literature, co-reference often involves linking specialized entity mentions. For instance, a protein described as \emph{“the HSPB2 gene product”} could subsequently be referenced by the phrase \emph{“this protein”}, with both mentions referring to the same biological entity \cite{Li2021}. 


Gan et al. (2024) evaluated the effectiveness of LLMs, including GPT-3.5, GPT-4, and LLAMA2, for coreference resolution using zero-shot prompting \cite{Gan2024}. GPT-4 achieved a high accuracy of 94\% on the WINOGRANDE benchmark \cite{Sakaguchi2020}, prompting further investigation into potential dataset exposure during training. To address this, Gan et al. created 44 handcrafted examples, where GPT-4 still achieved an accuracy of 93.2\%. However, on the formal CRAC datasets, initial automatic evaluations showed lower performance across all LLMs. Subsequent human annotation revealed that this discrepancy arose due to variability in LLM-generated outputs rather than true comprehension errors. After review, LLMs often outperformed baseline methods, demonstrating robust co-reference resolution capabilities despite occasional output inconsistencies.

Thus, in order for a co-reference system to have the best and most versatile performance in the biomedical field the system must have a deep understanding of the biomedical domain. GPT-4 has demonstrated remarkable proficiency across diverse language tasks, including a deep understanding of medical content as evidenced by its superior performance on USMLE exams and other medical benchmarks \cite{Nori2023}. Its extensive training on varied corpora, combined with robust zero-shot and few-shot learning capabilities through prompt engineering, makes GPT-4 a highly versatile candidate for IE systems where resolving co-references across both biomedical and general texts is critical. The following diagram shows this process on the extracted propositions.

\begin{figure}
    \includegraphics[width=\linewidth]{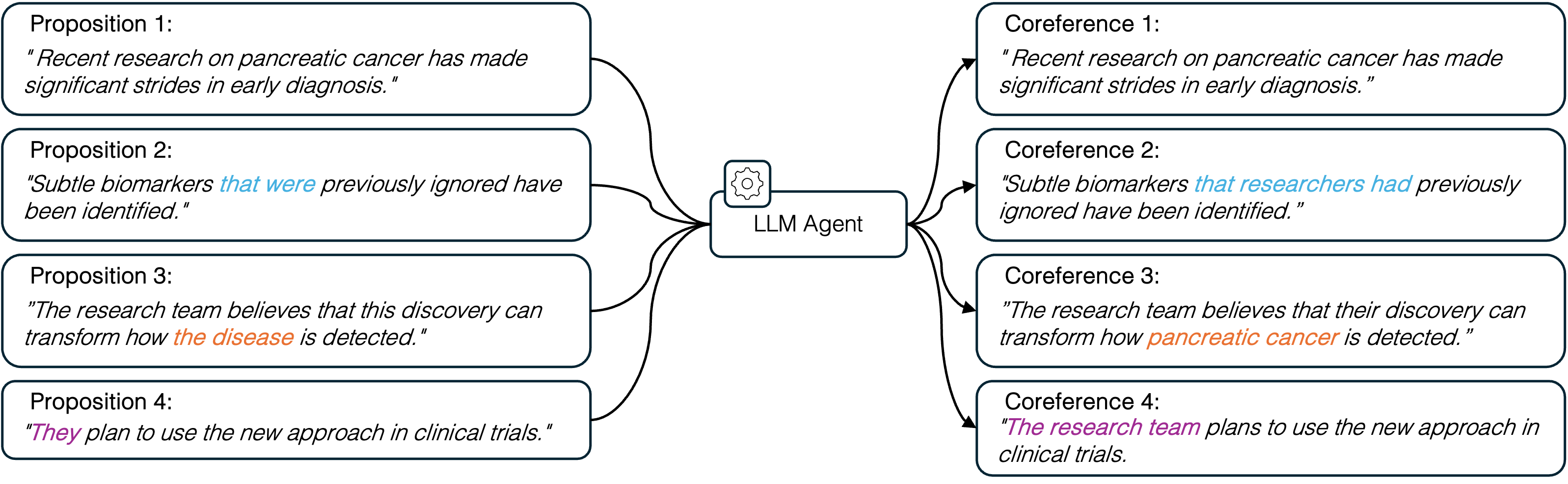}
    \caption{Example of Co-reference System.}
    \label{fig:coreference_example}
\end{figure}



\subsection{Information extraction}

It is clear that there are benefits of using an ontology for information extraction because it minimises the risk of extracting false relations, it forces valid relations and improves the ability of the extracted knowledge to be mapped to existing knowledge bases. However, because ontologies are restrictive and don't allow novel relations, their ability to map a domain of knowledge is limited. Open domain extraction has a better chance of extracting novel relations but has weaknesses when conforming to existing knowledge bases and being easily validated by researchers. Thus, a novel blended approach will be used. Although transformer models like BioBERT, BioBert and PubMedBERT show a state-of-the-art in relation extraction \cite{Vaswani2017,Lee2020,Beltagy2019,Gu2021}, they are restricted to a predefined ontology. Thus, LLMs were selected for relation extraction due to their versatility and comprehensive domain knowledge. Their deep understanding of the biomedical field enables them to infer relevant ontological categories for extracted entities and relationships, even when employing open-domain extraction methods \cite{Nori2023}. 

Evaluating the following co-reference sentence from Figure \ref{fig:coreference_example} for information extraction:
\begin{quote}
``Subtle biomarkers that researchers had previously ignored have been identified.''
\end{quote}

A conventional system might extract the following subject-relation-object triple:

\[
\texttt{(subtle biomarkers, have been identified by, researchers)}
\]

Although this triple captures the basic relationship, it lacks the additional context required to fully represent the meaning of the original sentence. Without context, the extracted triple does not convey critical information—such as that researchers had previously ignored these biomarkers.

In an enhanced extraction approach, each triple is augmented with an extra value (context) and include entity and predicate types to provide a richer, more complete representation -- thereby becoming a quadruple. The context in this scenario is defined to be the reason/justification of why the relationship was extracted. The following representation illustrates this enhanced extraction on the previous example:

  \begin{itemize}
    \item \textbf{Subject:} Subtle biomarkers (\textit{Types: Biomarker, Biological Entity})
    \item \textbf{Predicate:} have been identified by (\textit{Types: Action, Discovery})
    \item \textbf{Object:} Researchers (\textit{Types: Researcher, Professional, Human})
    \item \textbf{Context:} Researchers have discovered previously overlooked subtle biomarkers, highlighting advancements in the identification of biological indicators.
  \end{itemize}

This representation shows that by having the LLM add a context variable (the ``reason'') and including types for entities and relationships, it demonstrates a more comprehensive and faithful representation of the original proposition. The additional context ensures that the nuanced meaning—such as that the biomarkers were previously overlooked—is preserved, making the extracted information more interpretable for downstream applications.

\subsection{Round-trip Validation}

\begin{figure}[ht]
  \includegraphics[width=0.9\textwidth]{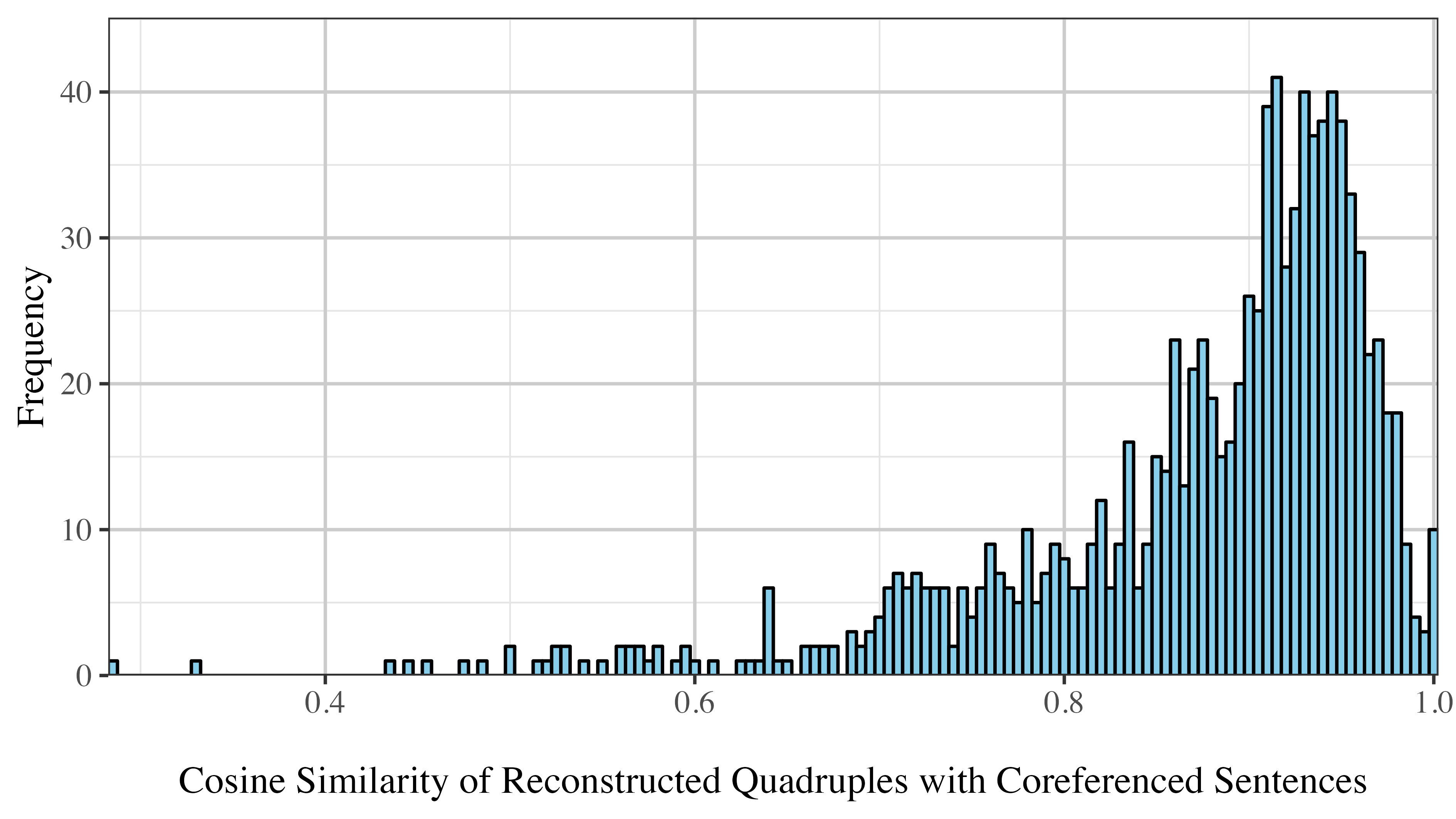}
  \caption{Distribution of cosine similarities between the reconstructed quadruples and co-referenced sentences in 44 abstracts (0 is no similarity and 1.0 is 100\% similarity).}
  \label{fig:original_quadruples_hist}
\end{figure}

Round-trip validation is the process of converting data from one format into another format and then back again, ensuring that the final output matches the original input with no loss or alteration. It is a form of bidirectional transformation commonly used in file format conversions, language translation systems and data serialization/deserialization \cite{ref_fowler2002}. In this specific use case, the system has access to the co-referenced sentence and the extracted quadruple. This allows for an LLM to reconstruct the co-referenced sentence from the quadruple and then compare the similarity of the reconstructed sentence with the original co-referenced sentence to quantify their semantic equivalence. This conversion is important to mitigate the impact of an LLM hallucinating an incorrect quadruple. Sentence embedding is used to compare the semantic similarity of the respective sentences. Both sentences are encoded into vectors and the cosine similarity of these vectors is calculated to produce a value between -1 and 1. Where a value closer to 1 means the embeddings are identical, a value closer to 0 shows no similarity and a value closer to -1 shows perfect dissimilarity. The model chosen for this task is Paraphrase-MiniLM-L6-v2 which is a lightweight transformer-based sentence embedding model \cite{ref_reimers2019}. The model was chosen for its strong performance on the Semantic Textual Similarity Benchmark with a spearman correlation score of 0.86 \cite{ref_reimers2019}. The results of running this semantic similarity on all reconstructed quadruples in all 44 abstracts produced a right skewed distribution centered around the median value 0.908 with a 25\% quartile at 0.837 and a 75\% quartile at 0.942. The distribution of these values can be seen in Figure \ref{fig:original_quadruples_hist}.

\begin{figure}[ht]
  \includegraphics[width=0.9\textwidth]{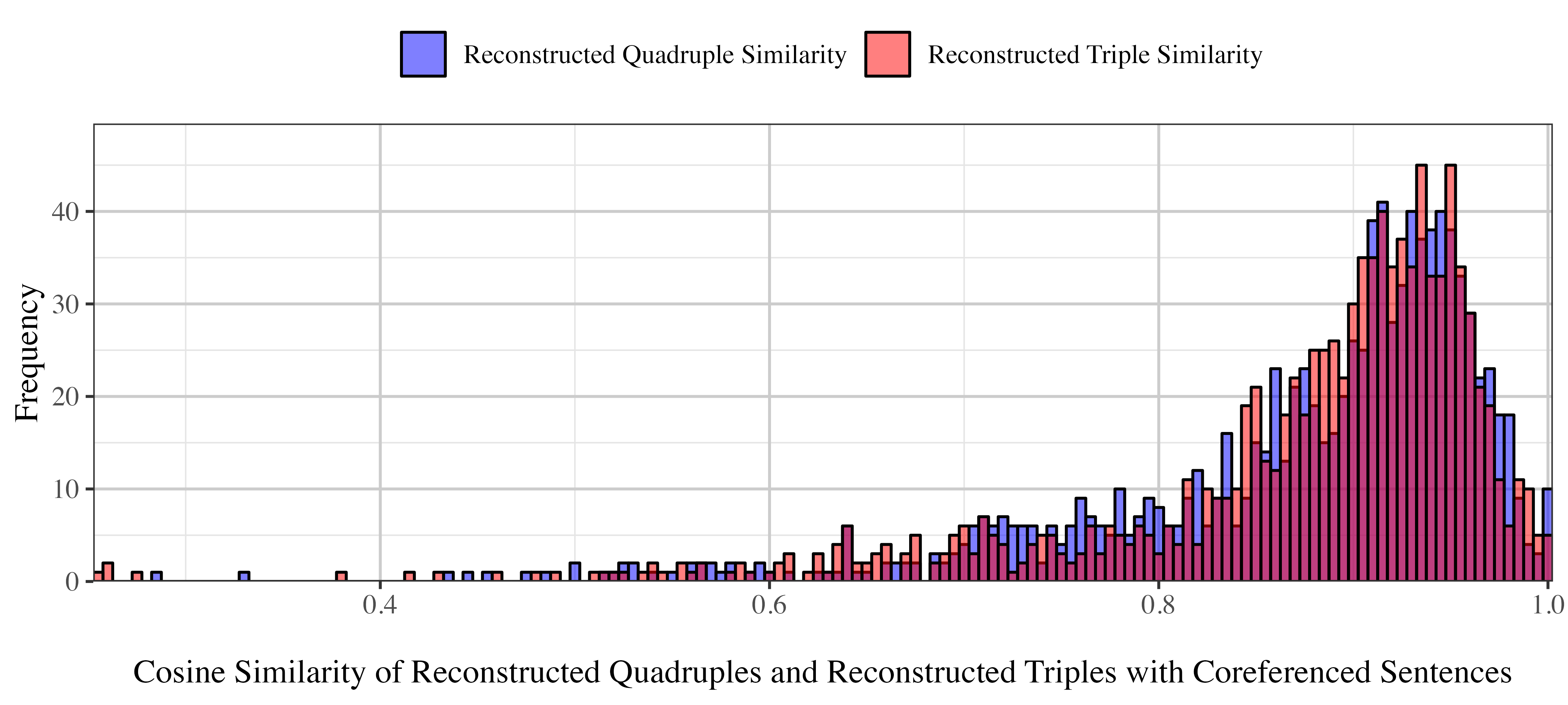}
  \caption{Distribution of cosine similarities between the reconstructed quadruples and the reconstructed triples with the co-referenced sentences in 44 abstracts.}
  \label{fig:dual_hist}
\end{figure}

In order to prove the effectiveness of adding a context variable, an LLM was used to reconstruct the extracted triple (the same quadruple without the context variable) into a natural language sentence and compare the similarity of this sentence with the co-referenced sentence against that of the reconstructed quadruple. The overall results showed an almost even semantic similarity over the 44 abstracts. The triple similarity showed a median of 0.905 with a 25\% quartile at 0.848 and a 75\% quartile at 0.940. The comparative distribution can be seen in Figure \ref{fig:dual_hist}.

\begin{figure}[ht]
  \includegraphics[width=0.9\textwidth]{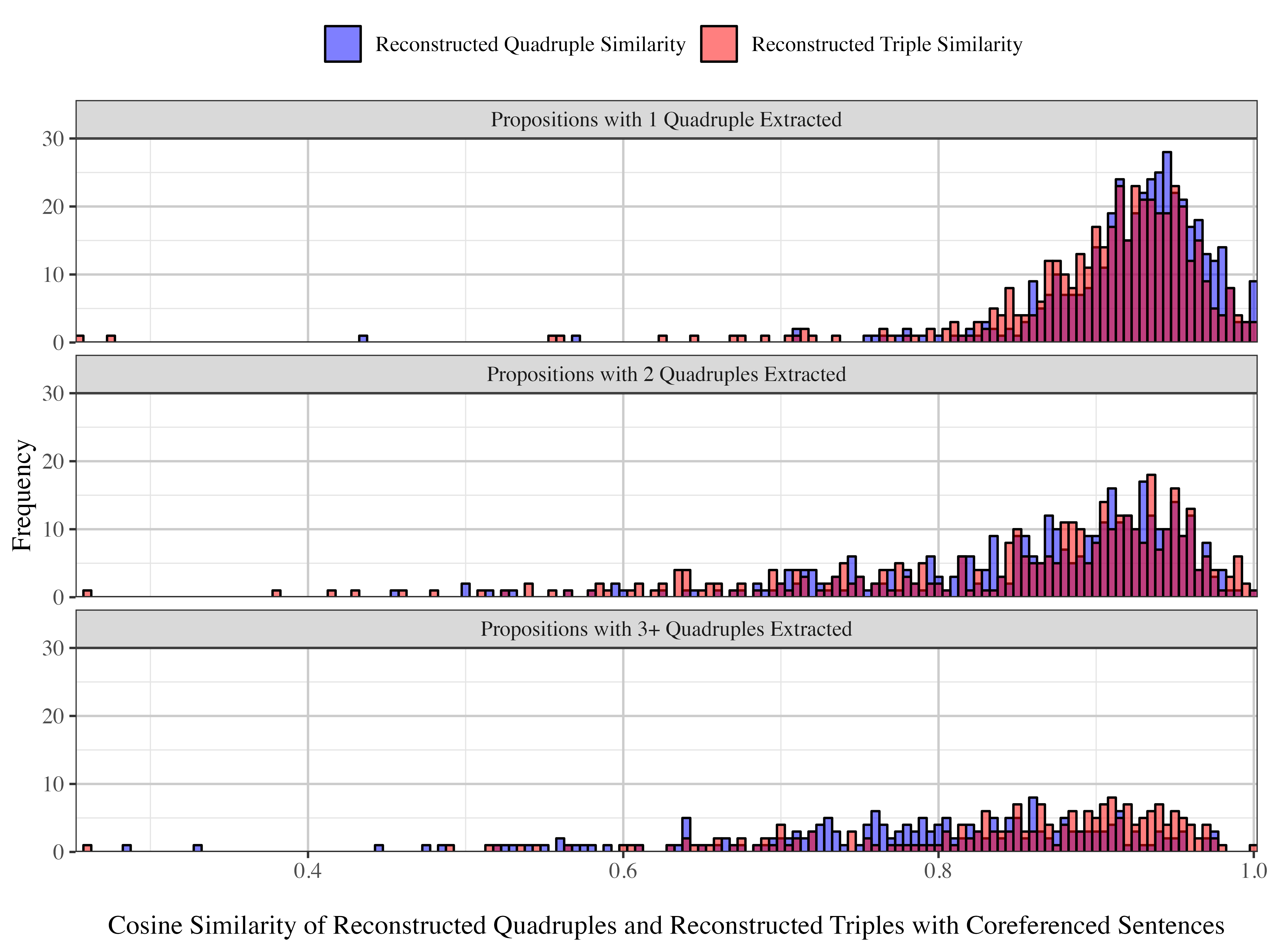}
  \caption{Distribution of cosine similarity between the reconstructed quadruples and triples for the number of quadruples extracted from a proposition.}
  \label{fig:stacked_comparison}
\end{figure}

However, when examining the similarity based on the number of quadruples extracted from a co-reference sentence. The results, given in Figure \ref{fig:stacked_comparison}, show that the reconstructed quadruples outperformed the reconstructed triples on sentences where 1 or 2 quadruples were extracted but performed far worse on sentences where 3 or more quadruples were extracted. 

This trend is further supported by the central tendency metrics in Table \ref{tab:similarity_metrics}. When only one quadruple was extracted, the reconstructed quadruples had a higher median similarity (0.933) compared to the reconstructed triples (0.921). Similarly, in cases where two quadruples were extracted, both approaches performed nearly identically, with medians of 0.892 and 0.890, respectively. However, when three or more quadruples were extracted, the reconstructed triples outperformed the reconstructed quadruples, achieving a far higher median similarity (0.879 vs. 0.804). When explicitly looking at the cases where 3 or more quadruples were extracted, almost all cases contained a list of some sort. For example, ``Key modifiable risk factors include cigarette smoking, obesity, diabetes, and alcohol intake.'', where each quadruple will contain a single aspect of this list and the added context further differentiates the reconstructed quadruple from the other aspects in the list. These lists occurred because co-reference resolution took place after proposition chunking, positioning lists in place of general entities thus producing propositions that were not in their most finite form. As such, these cases should not be considered when evaluating the performance of reconstruction.

\begin{table}[H]
\caption{Comparison of central tendency metrics between reconstructed quadruples and triples for the number of quadruples extracted from a proposition.}
\label{tab:similarity_metrics}
\centering
\begin{tabular}{|l|c|c|c|c|c|c|}
\hline
\textbf{Metric} & \multicolumn{2}{c|}{\thead{\textbf{1 Quadruple}\\\textbf{Extracted}}} & \multicolumn{2}{c|}{\thead{\textbf{2 Quadruples}\\\textbf{Extracted}}} & \multicolumn{2}{c|}{\thead{\textbf{3+ Quadruples}\\\textbf{Extracted}}} \\ \hline
\thead{Percentage\\of Dataset} & \multicolumn{2}{c|}{42.96\%} & \multicolumn{2}{c|}{36.76\%} & \multicolumn{2}{c|}{20.28\%} \\ \hline
 & \thead{\textbf{Original}} & \thead{\textbf{Triple}} & \thead{\textbf{Original}} & \thead{\textbf{Triple}} & \thead{\textbf{Original}} & \thead{\textbf{Triple}} \\ \hline
Median & 0.933 & 0.921 & 0.892 & 0.890 & 0.804 & 0.879 \\ \hline
25th Percentile & 0.905 & 0.885 & 0.820 & 0.795 & 0.727 & 0.807 \\ \hline
75th Percentile & 0.955 & 0.947 & 0.932 & 0.935 & 0.877 & 0.923 \\ \hline
\end{tabular}
\end{table}

\begin{figure}[H]
  \centering
  \includegraphics[width=0.8\textwidth]{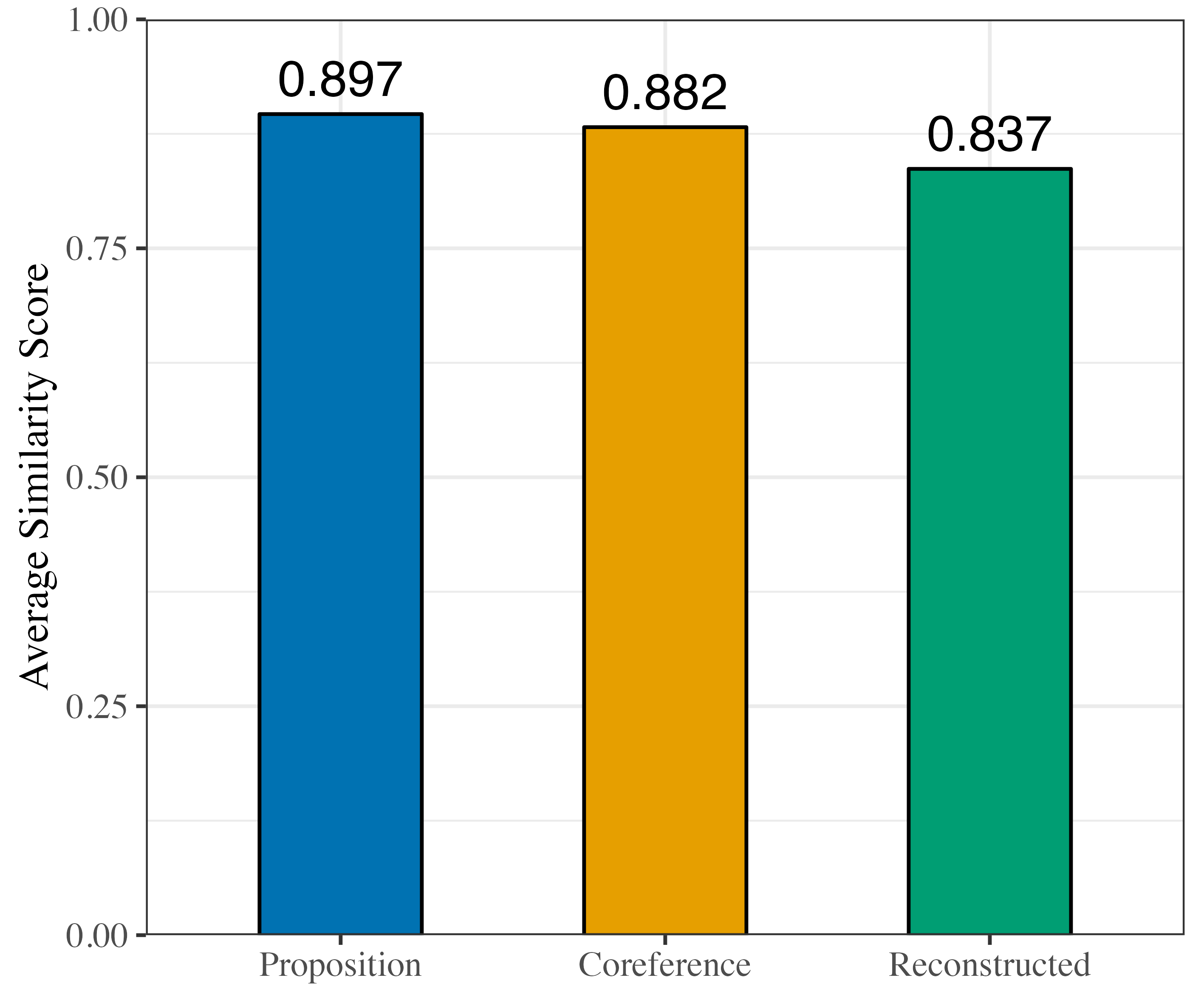}
  \caption{Average cosine similarity between combined propositions, co-references and reconstructed quadruples and the original abstracts.}
  \label{fig:paragraph_level_comparison}
\end{figure}

Throughout the extraction process, the proposition sentences, co-reference sentences and reconstructed sentences are stored to produce an audit trail for the extraction process for each individual abstract. This allows for all of the respective proposition sentences, co-reference sentences and reconstructed sentences to be combined, forming their own copies of the original abstract. These paragraphs, being within the token limit of the embedding model \cite{ref_reimers2019}, allowed for a high-level round-trip validation with the original abstract. The results were then averaged over all 44 abstracts to show the information loss through the system. The results can be seen in Figure \ref{fig:paragraph_level_comparison}.

\subsection{Inferred Quadruples}

A limitation with the extracted KG for each chunk was often found in the nature of entities. Because an entity or node is a string unit, in order for each triple to be able to connect, they must have the exact same node. Often, this leads to the resulting knowledge graph of a text chunk having clusters of nodes that do not connect with each other. Although this may mean that the meanings of these nodes do not align, downstream tasks will be effected by the disjunction of information when graph traversal is necessary \cite{ref_togon}. 

Therefore, a process is designed using an LLM to connect the clusters of connecting quadruples. The LLM will take in clusters of quadruples as well as the original abstract and output inferred quadruples where the relationship of the inferred quadruple is used to connect two clusters and the context of the quadruple is used as a justification for why that relationship was inferred. It was found that giving the LLM all of the clusters and the abstract in one sitting produced few inferred quadruples that actually connected the clusters. Therefore, given $n$ clusters, the two largest clusters are given to the LLM agent together with the abstract to infer new quadruples. The inferred new quadruples are added to the total quadruples and the quadruples are re-clustered (producing $n-1$ clusters). This process is repeated $n-1$ times until there is a fully connected graph. The figure below shows a knowledge graph before and after inferring quadruples. 

\begin{figure}[H]
  \centering
  \includegraphics[width=\textwidth]{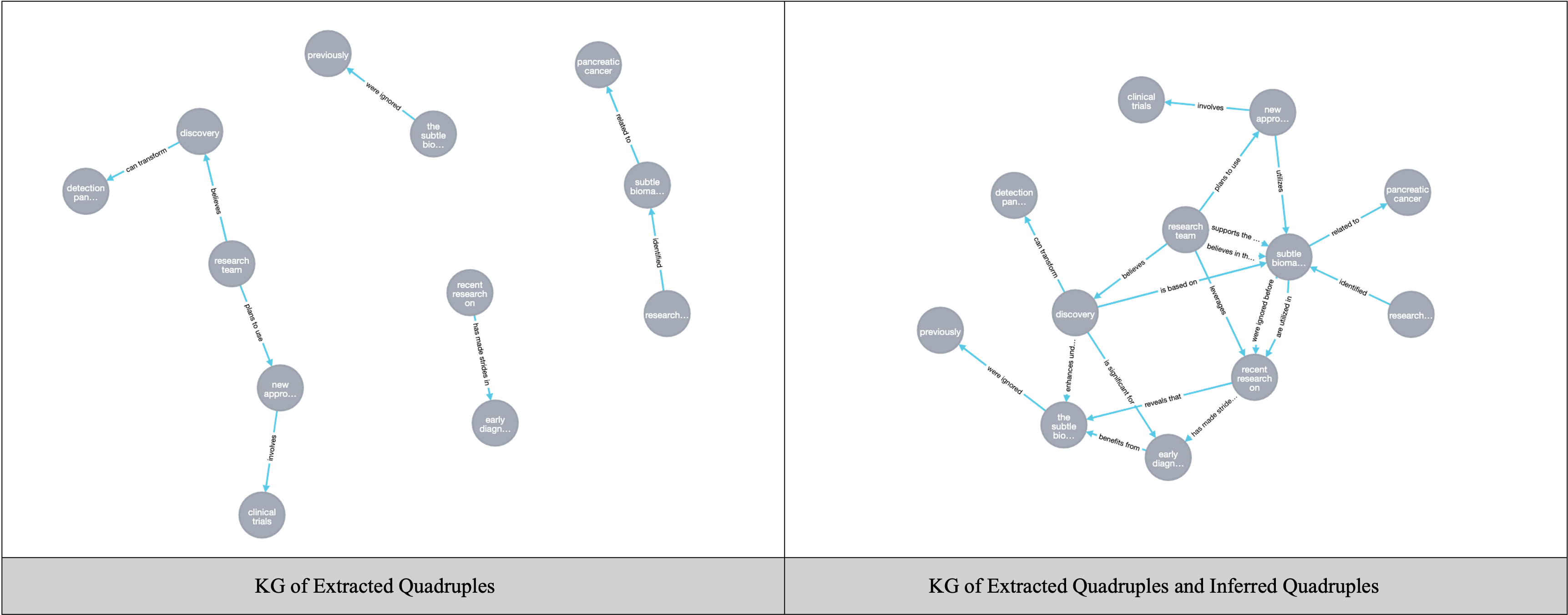}
  \caption{Knowledge graph comparison showing the effect of inferred quadruples on clusters of nodes.}
  \label{fig:inferred}
\end{figure}

\section{Limitations and Validity}

While the system demonstrates strong performance in extracting structured biomedical knowledge, several limitations and validity concerns must be acknowledged. Internal validity is maintained through round-trip validation to reduce hallucinations in LLM-based extraction, though reliance on a single model (GPT-4) introduces potential biases in sentence decomposition and relationship inference. External validity is challenged by the focus of the study on 44 PubMed abstracts, limiting generalisability to broader biomedical texts. Key threats to validity include dataset size constraints, LLM-generated misleading relationships, and inconsistencies in biomedical nomenclature affecting entity linking and ontology labelling. Additionally, the system faces two primary limitations: (i) computational costs, as LLM-based extraction is resource-intensive and expensive; (ii) context window constraints, which impact the processing of long input texts.


\section{Conclusion}

A context-aware IE pipeline has been presented to address the challenge of extracting structured biomedical knowledge from unstructured text. Through the use of LLMs, PubMed abstracts are decomposed into proposition sentences, from which KG triples are extracted and enhanced into quadruples through open domain and ontology-based methodologies. The inclusion of a context variable to the extracted triple improves the interpretability of extracted information, with validation showing a cosine similarity of 0.874 between reconstructed and original sentences. Further, inferred relationships enhance connectivity within the knowledge graph, facilitating a more comprehensive representation of biomedical knowledge.

%
%
%
%

\end{document}